\documentclass[letterpaper]{article} % DO NOT CHANGE THIS
\usepackage{aaai2026}  % DO NOT CHANGE THIS
\usepackage{times}  % DO NOT CHANGE THIS
\usepackage{helvet}  % DO NOT CHANGE THIS
\usepackage{courier}  % DO NOT CHANGE THIS
\usepackage[hyphens]{url}  % DO NOT CHANGE THIS
\usepackage{graphicx} % DO NOT CHANGE THIS
\usepackage{natbib}  % DO NOT CHANGE THIS AND DO NOT ADD ANY OPTIONS TO IT
\usepackage{caption} % DO NOT CHANGE THIS AND DO NOT ADD ANY OPTIONS TO IT
\usepackage{color}
\usepackage{algorithm}
\usepackage{algorithmic}
\usepackage{newfloat}
\usepackage{listings}
\usepackage{multirow}
\usepackage[table]{xcolor}
\usepackage{booktabs} % 提供 \Xhline
\usepackage{amssymb}
\usepackage{amsmath}
\usepackage{makecell}
\usepackage{caption}
\usepackage{tabularray}
\usepackage{arydshln} 
\usepackage{pifont} % 在导言区添加

\DeclareCaptionStyle{ruled}{labelfont=normalfont,labelsep=colon,strut=off} % DO NOT CHANGE THIS
\floatstyle{ruled}
\newfloat{listing}{tb}{lst}{}
\floatname{listing}{Listing}
\title{PIF-Net: Ill-Posed Prior Guided Multispectral and Hyperspectral Image Fusion via Invertible Mamba and Fusion-Aware LoRA}

\author{
    Baisong Li\textsuperscript{\rm 1,2}, 
    Xingwang Wang\textsuperscript{\rm 1,2}\thanks{Corresponding Author}, 
    Haixiao Xu\textsuperscript{\rm 1,2}
}
\affiliations{
    \textsuperscript{\rm 1}College of Computer Science and Technology, Jilin University\\
    \textsuperscript{\rm 2}Key Laboratory of Symbolic Computation and Knowledge Engineering of Ministry of Education, Jilin University\\
    lbs23@mails.jlu.edu.cn, \{xww, haixiao\}@jlu.edu.cn
}
\begin{document}

\maketitle
\begin{abstract}
The goal of multispectral and hyperspectral image fusion (MHIF) is to generate high-quality images that simultaneously possess rich spectral information and fine spatial details. However, due to the inherent trade-off between spectral and spatial information and the limited availability of observations, this task is fundamentally ill-posed. Previous studies have not effectively addressed the ill-posed nature caused by data misalignment. To tackle this challenge, we propose a fusion framework named PIF-Net, which explicitly incorporates ill-posed priors to effectively fuse multispectral images and hyperspectral images. To balance global spectral modeling with computational efficiency, we design a method based on an invertible Mamba architecture that maintains information consistency during feature transformation and fusion, ensuring stable gradient flow and process reversibility.
Furthermore, we introduce a novel fusion module called the \textit{Fusion-Aware Low-Rank Adaptation} module, which dynamically calibrates spectral and spatial features while keeping the model lightweight. Extensive experiments on multiple benchmark datasets demonstrate that PIF-Net achieves significantly better image restoration performance than current state-of-the-art methods while maintaining model efficiency.
% Code at \textcolor{magenta}{https://anonymous.4open.science/r/PIF-Net-B8BF}.
\end{abstract}
% \noindent \textbf{Code}: https://anonymous.4open.science/r/PIF-Net-B8BF

\begin{figure}[!htb]
    \centering
    \includegraphics[width=1.0\linewidth]{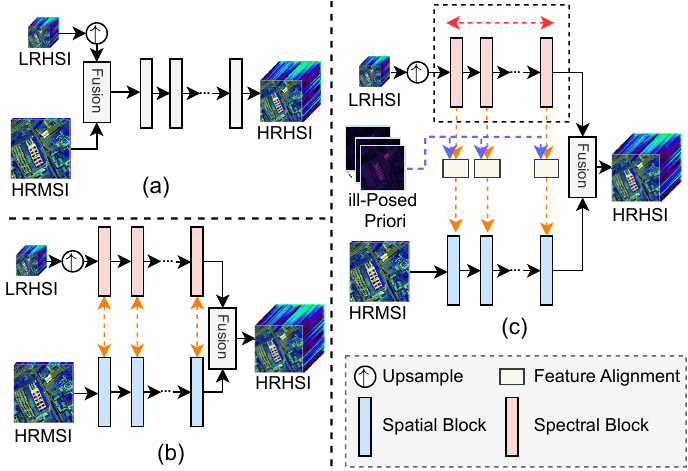}
    \caption{Illustration of three typical hyperspectral image fusion frameworks: (a) Single-branch framework: relies on a single pathway and is prone to alignment errors; (b) Dual-branch framework: models spectral and spatial features separately but lacks effective spatial transformation and collaborative mechanisms; (c) The proposed PIF-Net model: integrates invertible  state-space modeling with fusion-aware LoRA, enabling bidirectional information flow in the frequency domain and robust cross-modal alignment.}
    \label{fig:top}
\end{figure}

\section{Introduction}
Multispectral and Hyperspectral Image Fusion (MHIF) aims to effectively combine the rich spatial details from multispectral images (MSI) with the continuous and abundant spectral information from hyperspectral images (HSI), generating high-quality images that possess both high spatial resolution and high spectral resolution. By integrating data from these two modalities, MHIF seeks to overcome the limitations of single sensors in either spatial or spectral resolution, enhancing spectral fidelity and spatial detail representation in the fused images. This, in turn, better supports vision-based downstream applications such as image classification~\cite{class_01,class_02,class_03}, object detection~\cite{od_02,od_01,od_03}, and other practical tasks~\cite{agriculture,diagnosis,application}.

Most fusion methods based on single-branch (as Figure~\ref{fig:top}(a))~\cite{Fusformer,3DT-Net,PSRT} or dual-branch (as Figure~\ref{fig:top}(b))~\cite{U2Net,HSRnet} architectures, despite showing improvements over traditional model-driven approaches~\cite{low_rank01,Brovey,GSA}, still suffer from two fundamental limitations:
\textit{
(1) Significant differences in spectral characteristics, spatial resolution, and data distribution between MSIs and HSIs result in severe information misalignment and a substantial modality gap between the two sources. This constitutes a highly ill-posed problem that renders unified and effective feature modeling particularly challenging, thereby constraining further advances in fusion performance.
(2) Existing fusion methods commonly fail to ensure reversible information transfer, leading to inevitable information loss during the fusion process. Such irreversible loss directly impairs spectral fidelity and the high-quality recovery of spatial details.
}

To address the ill-posedness inherent in image restoration or image fusion, researchers have conducted extensive studies and explorations. For example, the \textit{Structure-Aware Deep Network}~\cite{structure_02,structure_01} enhances the model’s ability to capture geometric structures in images by introducing a structure-aware module, thereby significantly improving the visual quality of super-resolved images. The adoption of generative adversarial training~\cite{gan,pan-gan,esrgan} further facilitates the generation of realistic high-resolution images with complex structures. Denoising diffusion models~\cite{pan-diffusion,sr3} have demonstrated excellent performance in detail restoration and noise suppression, producing more natural and realistic images while effectively mitigating the over-smoothing caused by the ill-posed nature of the problem. \textit{However, these methods typically rely on general structural priors and mainly focus on low-level fixed patterns, making it difficult to capture the semantic shifts introduced during fusion.}

Furthermore, the introduction of invertible neural networks (INNs)~\cite{2014NICE,xiao2020invertible} offers a novel perspective for image restoration tasks. Unlike traditional unidirectional mapping, invertible networks learn bidirectionally consistent mappings, establishing a reversible information flow between image degradation and restoration~\cite{Gflow,lugmayr2020srflow}, thereby theoretically preventing information loss. This mechanism not only helps alleviate the ill-posedness in the restoration process but also significantly enhances the model’s capability to transfer information across different modalities, making it particularly suitable for the complex cross-modal modeling demands in multi-source image fusion. \textit{Nevertheless, effectively leveraging the invertible structure to address the issue of information misalignment caused by modality inconsistencies remains a significant challenge.}

To address the above issues, this paper proposes an advanced fusion framework guided by ill-posed priors, as shown in Figure~\ref{fig:top}(c), called PIF-Net. This framework ingeniously integrates the invertible visual state-space model Mamba with a lightweight \textit{Fusion-Aware Multi-Head Low-Rank Adaptation} (FAM-LoRA) module , achieving high-fidelity and efficient multimodal image fusion. PIF-Net consists of two complementary branches: the frequency domain branch precisely separates high- and low-frequency components using a two-dimensional HWT, combined with the invertible Mamba module to enhance bidirectional information flow, ensuring lossless extraction of global spatial features and structural fidelity; the spatial domain branch leverages low-frequency global guidance and the lightweight LoRA module to finely extract rich spatial texture information, incorporating \textit{Large-Kernel Attention} (LKA)~\cite{lka1,lka2,lka3} and \textit{Squeeze-and-Excitation} (SE)~\cite{SENet} mechanisms to significantly improve cross-modal semantic modeling. Furthermore, PIF-Net introduces guided feature consistency loss and volumetric-aware loss to strengthen the consistency between the dual-modal features, ensuring high precision in modality alignment and feature fusion with semantic coherence. Extensive experimental results demonstrate that PIF-Net outperforms current state-of-the-art methods across multiple hyperspectral datasets, achieving outstanding fusion quality and computational efficiency. 

The main contributions of this work are summarized as follows:
\begin{itemize}
\item We propose PIF-Net, a novel fusion network combining invertible Mamba and fusion-aware LoRA, effectively tackling modality mismatch and ill-posedness in hyperspectral–multispectral image fusion for high-fidelity, efficient results.

\item We design a dual-branch architecture: the frequency branch combines Haar wavelet decomposition and invertible Mamba to enable bidirectional information flow and structural preservation in the frequency domain; the spatial branch uses low-frequency guidance and a lightweight LoRA module to extract semantically rich and stable spatial features.

\item  Extensive experiments on several public hyperspectral datasets demonstrate that PIF-Net significantly outperforms existing state-of-the-art (SOTA) methods in terms of fusion quality and computational efficiency.
\end{itemize}

\begin{figure*}[htb]
    \centering
    \includegraphics[width=1.0\linewidth]{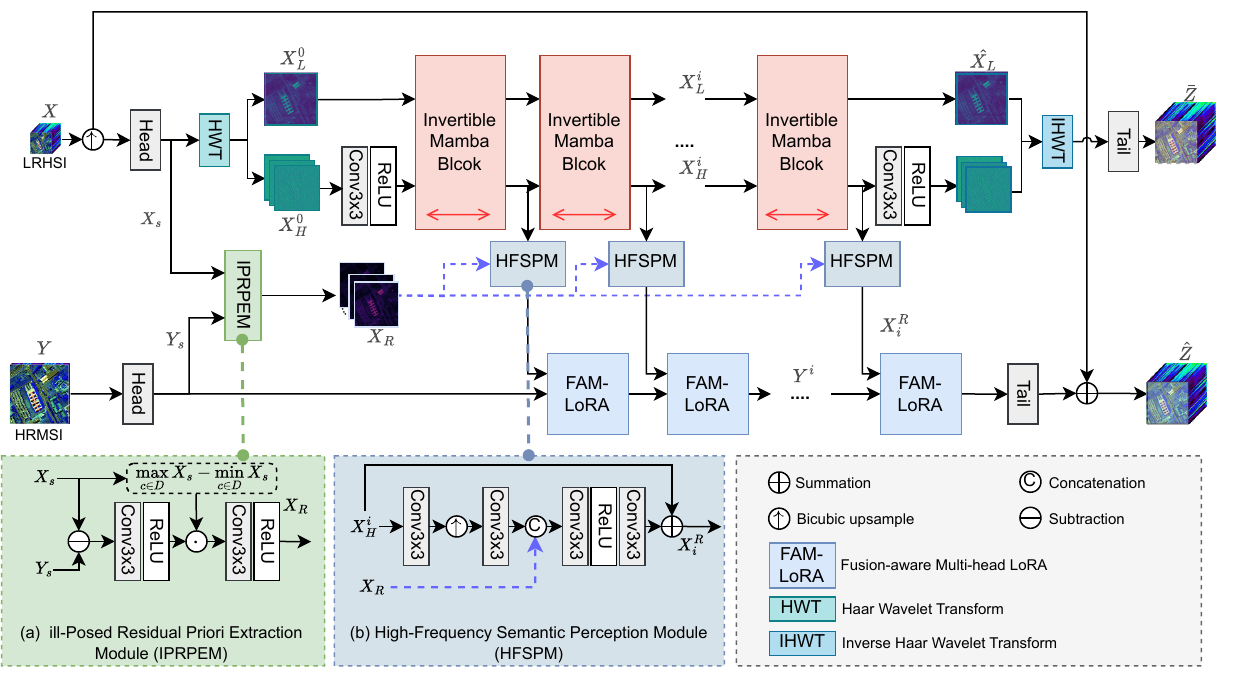}
    \caption{Overview of the proposed PIF-Net. The spectral branch utilizes \textit{Invertible Mamba Blocks} to enable bidirectional flow of low-frequency and high-frequency information; the spatial branch effectively extracts rich spatial texture features under the guidance of ill-posed residual priors and high-frequency spatial features.}
    \label{fig:overall}
\end{figure*}

\section{Related Work}

\subsection{Review of Existing Methods in MHIF}
In the field of multispectral and hyperspectral image fusion (MHIF), traditional methods such as pansharpening~\cite{pan15,GSA} and multiscale decomposition~\cite{deco} offer good interpretability but heavily rely on prior knowledge, making them prone to failure in complex scenarios. Bayesian approaches~\cite{18,19} can effectively model the fusion process but require careful parameter tuning.
In recent years, deep learning methods have significantly advanced MHIF. Early convolutional neural network (CNN)-based models, such as MHFnet~\cite{MHFnet} and HSRnet~\cite{HSRnet}, perform well under general conditions but struggle to capture long-range dependencies, leading to degraded performance in complex cases. To address this, Transformer-based Fusformer~\cite{Fusformer} was introduced, which greatly improves fusion quality but at a higher computational cost. Other approaches like PSRT~\cite{PSRT}, U2Net~\cite{U2Net}, and 3DT-Net~\cite{3DT-Net} bring innovations in architecture design but still face limitations in fusion capability, high training costs, or increased model complexity. SMGU-Net~\cite{SMGU-Net} attempts to combine traditional methods with deep learning to enhance cross-modal interactions, yet there remains substantial room for improvement in modeling complex nonlinear cross-modal relationships.

\subsection{Invertible Neural Networks}
Invertible Neural Networks (INNs) are a key component of normalizing flow models. Since NICE~\cite{2014NICE} introduced additive coupling layers, methods like RealNVP~\cite{RealNVP} and Glow~\cite{Gflow} have improved INNs, enabling efficient high-resolution image generation.
Beyond generative tasks, INNs can losslessly preserve features and save memory in classification. They also show broad potential in image colorization~\cite{ardizzone2019guided}, steganography~\cite{zhang2019steganogan}, image rescaling, and super-resolution~\cite{xiao2020invertible,lugmayr2020srflow}.
However, INNs remain underexplored in image fusion. For example, CDDFuse~\cite{zhao2023cddfuse} uses convolution-based invertible modules for detail extraction but fails to fully exploit spatial guidance from auxiliary modalities to improve fusion quality.

\section{Proposed Method}

\subsection{Overall structure}
Given a low-resolution hyperspectral image (LRHSI) $\mathbf{X} \in \mathbb{R}^{h \times w \times C}$ and a high-resolution multispectral image (HRMSI) $\mathbf{Y} \in \mathbb{R}^{H \times W \times c}$, PIF-Net employs a fusion mapping function $M(\cdot)$ to generate a high-resolution hyperspectral image (HRHSI) $\mathbf{\hat{Z}} \in \mathbb{R}^{H \times W \times C}$, where the scale factor is $s = H/h$:
\begin{equation}
\mathbf{\hat{Z}} = M(\mathbf{X}, \mathbf{Y} \mid \theta).
\end{equation}

The spectral branch first upsamples the LRHSI using bicubic interpolation and extracts initial features through a “Conv $\rightarrow$ ReLU $\rightarrow$ Conv” \textit{Head} module. These features are then projected to dimension $D$ and decomposed via Haar wavelets into a low-frequency component $\mathbf{X}_L$ and a high-frequency component $\mathbf{X}_H$. The high-frequency component is reduced to $D$ channels through a convolutional layer, while the low-frequency component is modeled in the frequency domain by passing through $L$ invertible Mamba modules. Finally, the fused spectral reference image $\bar{\mathbf{Z}}$. is generated by applying the inverse wavelet transform followed by a \textit{Tail} module, which shares the same structure as the \textit{Head}. 

To enhance the global spatial guidance capability of the spatial branch, we design a lightweight \textit{High-Frequency Semantic Perception Module }(see Figure~\ref{fig:overall}(b)). This module effectively constructs a highly robust global spatial semantic reference by fusing high-frequency upsampled features with ill-posed residual information. In the spatial branch, FAM-LoRA serves as the spatial restoration module. The output of the $L$-th FAM-LoRA block is processed by the \textit{Tail} module to generate the final fused image $\hat{\mathbf{Z}}$.

\subsection{Ill-Posed Residual Prior Extraction Module}
The \textit{Residue Channel Prior} (RCP)~\cite{RCP_01,RCP_02}, as an effective ill-posed prior in image restoration, exhibits spatial invariance while preserving contextual information. Inspired by this, we propose a dynamic \textit{Ill-Posed Residual Prior Extraction Module} (IPRPEM) tailored for the MHIF task. This module is designed to adaptively capture spectral-domain ill-posed characteristics while preserving spatially shifted contextual features. Specifically, it extracts global ill-posed prior information from shallow feature differences in an adaptive manner, enhancing the discriminative capacity of subsequent feature fusion. The computation is formulated as:

\begin{equation}
\left\{
\begin{aligned}
C_R &= \max_{c \in D} \bar{\mathbf{X}}_s - \min_{c \in D} \bar{\mathbf{X}}_s, \\
\mathbf{X}_R &= \text{ReLU}\big(\text{Conv}(\mathbf{X}_s - \mathbf{Y}_s)\big) \cdot C_R, \\
\mathbf{X}_R &= \text{ReLU}\big(\text{Conv}(\mathbf{X}_R)\big),
\end{aligned}
\right.
\end{equation}

where $\bar{\mathbf{X}}_s$ denotes the mean feature across the channel dimension, and $D$ is the set of all channels. The modulation factor $C_R$ serves to amplify residual cues, guiding the network to focus on global structural variations, thereby enhancing the discriminability and robustness of the extracted prior features $\mathbf{X}_R$.
\begin{figure}[tb]
    \centering
    \includegraphics[width=1\linewidth,trim=0 0 0 0]{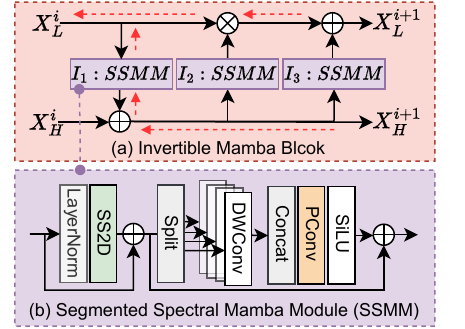}
    \caption{Illustration of the \textit{Invertible Mamba Block}. The block takes low-frequency features $\mathbf{X}^L_i$ and high-frequency features $\mathbf{X}^H_i$ as input, and employs an affine coupling mechanism built on lightweight \textit{Segmented Spectral Mamba Modules (SSMM)} to enable efficient bidirectional interaction and fusion of spectral information. \textit{Split} and \textit{Concat} denote channel-wise feature division and aggregation, respectively. SS2D refers to the \textit{2D State Space Module} in VMamba~\cite{liu2024vmamba}.}
    \label{fig:imamba}
\end{figure}

\subsection{Invertible Mamba Block}
CNNs~\cite{HSRnet,U2Net} struggle to capture long-range dependencies. Transformers~\cite{Fusformer} offer strong global modeling but suffer from quadratic complexity as feature size grows, reducing efficiency. To balance performance and cost, we adopt Mamba~\cite{liu2024vmamba,gu2023mamba}, a state-space model with linear complexity. Mamba captures global context efficiently and expressively. However, most Mamba-based models~\cite{guo2024mambair,guo2025mambairv2,li2025hsrmamba} still rely on irreversible designs, which cause feature ambiguity and information loss in image restoration. To solve this, we propose the \textit{Invertible Mamba Block}, which enables bidirectional interaction between low- and high-frequency spectral features (see Figure~\ref{fig:imamba}(a)). 

\textit{Invertible Mamba Block }uses coupling transformations from invertible neural networks. Each transformation unit is built with our \textit{Segmented Spectral Mamba Module} (SSMM), as shown in Figure~\ref{fig:imamba}(b). This design supports global spectral modeling and multi-scale feature extraction, improving representation quality and diversity. We split the input into high-frequency $\mathbf{X}_H^i$ and low-frequency $\mathbf{X}_L^i$ components and update them with the following invertible mapping:

\begin{equation}
\begin{aligned}
\mathbf{X}_H^{i+1} &= \mathbf{X}_H^i + I_1(\mathbf{X}_L^i), \\
\mathbf{X}_L^{i+1} &= \mathbf{X}_L^i \odot \exp\left[I_2(\mathbf{X}_H^i)\right] + I_3(\mathbf{X}_H^i).
\end{aligned}
\end{equation} 
Here, $\odot$ denotes element-wise multiplication. $I_1$, $I_2$, and $I_3$ are SSMM-based subnetworks designed to model nonlinear interactions between different spectral components. Thanks to the invertible architecture, the module enables efficient information fusion during the forward pass. Notably, to ensure the flexibility of invertible blocks in fusion tasks, \textit{we employ a gradient-guided invertibility strategy that heuristically facilitates backward interaction of spectral features. Instead of using traditional rescaling schemes~\cite{xiao2020invertible}, we design a tailored invertible loss function (see Eq.~\eqref{eq:loss})}.

\begin{table*}[tp]
\centering
\setlength\tabcolsep{3.2pt} 
\small
\resizebox{\textwidth}{!}{
\begin{tabular}{lccccccccccccc}
\toprule 
\multirow{2}{*}{Method}  &\multirow{2}{*}{Scale}  
&\multicolumn{4}{c}{Chikusei} 
&\multicolumn{4}{c}{Houston} 
&\multicolumn{4}{c}{PaviaU} \\
& &PSNR$\uparrow$&SSIM$\uparrow$  & SAM$\downarrow$ & ERGAS$\downarrow$  
& PSNR$\uparrow$ & SSIM$\uparrow$  & SAM$\downarrow$ & ERGAS$\downarrow$  
& PSNR$\uparrow$ & SSIM$\uparrow$  & SAM$\downarrow$ & ERGAS$\downarrow$  \\ 
\cmidrule(lr){1-2} \cmidrule(lr){3-6} \cmidrule(lr){7-10} \cmidrule(lr){11-14}
Brovey~\cite{Brovey}        &$\times$2&29.2704&0.8751&8.1522&11.0668 &  32.5312&0.8431&8.4829&5.7936&  27.8390&0.7958&6.2207&6.0319    \\ 
GSA~\cite{GSA}          &$\times$2&37.4039&0.9627&4.1328&7.0972 &   33.2998&0.8630&9.3221&5.4163&  28.7031&0.7945&6.6719&6.4978    \\ 
HSRnet~\cite{HSRnet}       &$\times$2&53.0324&0.9964&1.9876&1.4615 &   40.8710&0.9805&3.4652&1.9588&  40.9947&0.9685&3.6476&2.2422    \\ 
Fusformer~\cite{Fusformer}    &$\times$2&53.8104&0.9967&1.9414&1.4521 &   42.1020&0.9888&3.1807&1.9238&  41.8442&0.9727&3.2919&2.0589    \\
PSRT~\cite{PSRT}         &$\times$2&53.8821&0.9970&1.8994&1.3537 &   42.3937&0.9901&2.8675&1.8847&  41.9956&0.9790&2.2518&1.8971    \\ 
U2Net~\cite{U2Net}        &$\times$2&\underline{53.9657}&0.9974&1.8824&1.2972 &   42.7243&0.9920&2.7825&1.8308&  \underline{42.2347}&\underline{0.9813}&\underline{2.1596}&\underline{1.4162}    \\
3DT-Net~\cite{3DT-Net}      &$\times$2&53.9155&\underline{0.9988}&\underline{1.8222}&1.3098 &   \underline{43.1856}&\underline{0.9947}&\underline{2.5811}&\underline{1.7677}&  41.2224&0.9807&2.3080&1.6433    \\
SMGU-Net~\cite{SMGU-Net}     &$\times$2&52.7751&0.9985&1.8869&\underline{1.2842} &   43.1647&0.9946&2.6137&1.7902&  39.3881&0.9775&2.8193&1.8930   \\ 
PIF-Net       &$\times$2&\textbf{54.0236}&\textbf{0.9990}&\textbf{1.7962}&\textbf{1.1804}&  \textbf{43.4025}&\textbf{0.9963}&\textbf{2.4857}&\textbf{1.7498}&  \textbf{42.8673}&\textbf{0.9892}&\textbf{1.7289}&\textbf{1.1392}   \\
\midrule

Brovey~\cite{Brovey}        &$\times$4&27.5923&0.8325&6.2724&15.1083&   30.1662&0.7745&11.5137&7.9550& 25.5351&0.7143&8.8247&7.9284    \\ 
GSA~\cite{GSA}           &$\times$4&31.2757&0.8985&6.0018&10.3126&   30.9217&0.7605&12.3469&8.5070& 28.5528&0.6295&9.5707&10.0441   \\ 
HSRnet~\cite{HSRnet}       &$\times$4&49.3548&0.9901&2.4509&1.9242 &   40.4509&0.9835&3.8988&2.9216&  37.2608&0.9540&3.7748&3.1791    \\ 
Fusformer~\cite{Fusformer}    &$\times$4&50.1466&0.9909&2.3864&1.8927 &   41.1338&0.9895&3.4892&2.3436&  37.7936&0.9558&3.8403&2.9017    \\
PSRT~\cite{PSRT}         &$\times$4&50.5377&0.9919&2.2097&1.7832 &   41.5616&0.9882&3.2434&2.1695&  38.2794&0.9645&3.3579&2.4509    \\ 
U2Net~\cite{U2Net}        &$\times$4&50.5061&0.9929&2.1852&1.7317 &   41.7714&0.9889&3.1039&1.9978&  38.7681&0.9700&3.0176&2.2220    \\
3DT-Net~\cite{3DT-Net}      &$\times$4&48.1940&0.9961&2.4262&2.1546 &   42.6873&0.9931&3.0318&1.9952&  38.1864&0.9744&\underline{2.8405}&2.3481    \\
SMGU-Net~\cite{SMGU-Net}      &$\times$4&\underline{51.3382}&\underline{0.9981}&\underline{2.1080}&\underline{1.5762} &   \underline{42.9130}&\underline{0.9937}&\underline{2.8873}&\underline{1.8979}&  \underline{38.9201}&\underline{0.9761}&2.9800&\underline{1.9973}    \\

PIF-Net       &$\times$4&\textbf{51.6257}&\textbf{0.9983}&\textbf{2.0653}&\textbf{1.5401}&  \textbf{43.1128}&\textbf{0.9947}&\textbf{2.6452}&\textbf{1.7624}&  \textbf{39.8246}&\textbf{0.9845}&\textbf{2.4018}&\textbf{1.8126}    \\

\midrule
Brovey~\cite{Brovey}        &$\times$8&26.0978&0.7989&8.1166&18.1408&  27.0096&0.7217&16.0689&11.4050&   22.7245&0.6497&13.4666&10.7571  \\ 
GSA~\cite{GSA}           &$\times$8&28.1606&0.8609&8.0270&13.4385&  26.2180&0.6829&17.6147&11.9947&   21.2674&0.5245&14.1969&14.3429  \\ 
HSRnet~\cite{HSRnet}       &$\times$8&47.7262&0.9902&2.6291&2.3439 &  30.6384&0.9550&6.0210&5.9721&     30.1921&0.8956&6.0662&4.9901    \\ 
Fusformer~\cite{Fusformer}    &$\times$8&47.8910&0.9919&2.5806&2.2506 &  31.8052&0.9589&5.7651&5.4997&     30.4891&0.9011&5.9990&4.9715    \\ 
PSRT~\cite{PSRT}         &$\times$8&48.6457&0.9942&2.5291&2.0439 &  32.6384&0.9650&5.1151&5.1257&     31.7671&0.9189&5.6891&4.8965    \\ 
U2Net~\cite{U2Net}        &$\times$8&48.8911&0.9959&2.4006&1.9506 &  32.8052&0.9676&4.9910&4.9711&     \underline{31.9211}&0.9210&5.5451&4.6712    \\ 
3DT-Net~\cite{3DT-Net}      &$\times$8&44.9324&0.9936&2.9072&2.6482 &  34.1538&0.9688&5.0265&4.8583&     29.2755&0.9302&5.8975 &5.0295   \\
SMGU-Net~\cite{SMGU-Net}     &$\times$8&\underline{49.6582}&\underline{0.9975}&\underline{2.3260}&\underline{1.7872} &  \underline{34.2202}&\underline{0.9699}&\underline{4.6268}&\underline{4.8072}&     29.7848&\underline{0.9417}&\underline{5.4906}&\underline{4.4843}   \\ 

PIF-Net       &$\times$8&\textbf{50.0124}&\textbf{0.9979}&\textbf{2.2603}&\textbf{1.7328}&  \textbf{34.2989}&\textbf{0.9706}&\textbf{4.3902}&\textbf{4.7218}&     \textbf{32.2310}&\textbf{0.9521}&\textbf{5.4258}&\textbf{4.1726}  \\
\midrule
Best Value& -& $+\infty$&1&0&0 & $+\infty$&1&0&0 & $+\infty$&1&0&0 \\ 
\bottomrule
\end{tabular}
}
\caption{
Quantitative comparisons of different approaches were conducted on the Chikusei, Houston, and PaviaU test datasets.
The \textbf{best} results are in bold, and the \underline{second-best} results are underlined.}

\label{tab:main}
\end{table*}
\begin{figure}[tb]
    \centering
    \includegraphics[width=1\linewidth]{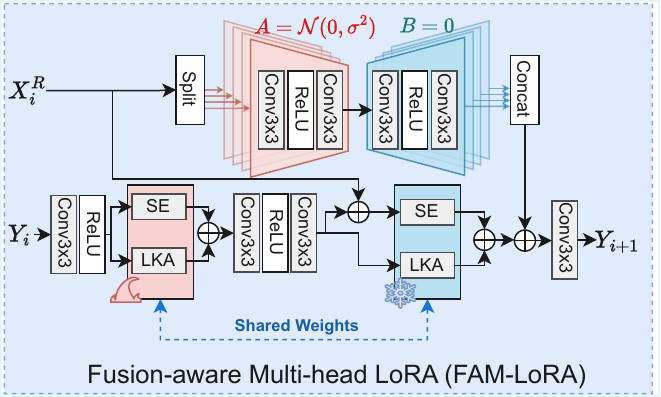}
   \caption{Illustration of the FAM-LoRA module that fuses the main spatial feature \(\mathbf{Y}_i\) and auxiliary guidance \(\mathbf{X}^R_i\) through channel transformation, LKA, SE attention, and multi-head LoRA, achieving efficient and accurate semantic fusion. The first part of LoRA's parameters is initialized with a standard Gaussian distribution, while the second part is initialized to zero. The operations \textit{Split} and \textit{Concat} refer to feature division and aggregation along the channel dimension. }
    \label{fig:lora}
\end{figure}

\subsection{Fusion-aware Multi-head LoRA}
Inspired by the success of LoRA in large language models and vision tasks~\cite{hu2022lora,dora,vis_lora_01,vis_lora_02}, we propose a Fusion-Aware Multi-head Low-Rank Adaptation module (FAM-LoRA), as shown in Figure~\ref{fig:lora}, which enables efficient semantic interaction during feature fusion with minimal parameters and computational overhead. Specifically, FAM-LoRA takes the main spatial feature $\mathbf{Y}_i$ and auxiliary guidance feature $\mathbf{X}^R_i$ as inputs. $\mathbf{Y}_i$ first passes through four consecutive $1\times1$ convolutions for channel compression and transformation, followed by LKA and SE modules to enhance spatial context awareness and channel-wise selectivity. The gradients of LKA and SE are then frozen, and $\mathbf{X}^R_i$ is injected into the SE input to steer attention toward critical semantic regions, improving fusion accuracy. To capture multi-scale representations, the channel dimension is evenly split into four heads, each independently applying Low-Rank Adaptation.

\subsection{Loss Function}

To achieve efficient training and precise representation in PIF-Net, the loss function is composed of three parts, aiming to comprehensively enhance the fusion accuracy and feature consistency of the model, while ensuring the invertibility of the transformation process. The loss function is formulated as follows:

\small{
    \begin{equation}
    \label{eq:loss}
        \mathcal{L} = \underbrace{\| \hat{\mathbf{Z}} - \mathbf{Z} \|_1}_{\mathcal{L}_1} - \lambda_{\mathrm{inv}} \cdot \underbrace{\log \left| \det \left( \frac{\partial \bar{\mathbf{Z}}}{\partial \mathbf{X}} \right) \right|}_{\mathcal{L}_{inv}} + \lambda_{\mathrm{cos}} \cdot \underbrace{\left( 1 - \cos(\bar{\mathbf{Z}}, \hat{\mathbf{Z}}) \right)}_{\mathcal{L}_{cos}}
    \end{equation}
}

First, the $\ell_1$ fusion loss \(\mathcal{L}_1 = \| \hat{\mathbf{Z}} - \mathbf{Z} \|_1\) precisely constrains the difference between the predicted feature \(\hat{\mathbf{Z}}\) and the ground truth feature \(\mathbf{Z}\), effectively reducing fusion error. 

Second, the log-determinant of the Jacobian of the affine transformation,
\(\mathcal{L}_{inv} = \log \left| \det \left( \frac{\partial \bar{\mathbf{Z}}}{\partial \mathbf{X}} \right) \right|\), serves as an invertibility regularization term to ensure information preservation and numerical stability during the mapping process~\cite{Gflow,lugmayr2020srflow}, encouraging the model to learn stable and invertible feature transformations. Finally, the cosine similarity loss
\(\mathcal{L}_{cos} = 1 - \cos(\bar{\mathbf{Z}}, \hat{\mathbf{Z}})\) enhances angular consistency between different feature representations, improving semantic feature fusion.

Overall, the loss function balances fusion fidelity (\(\mathcal{L}_1\)), semantic consistency (\(\mathcal{L}_{cos}\)), and transformation invertibility (\(\mathcal{L}_{inv}\)) under multi-objective constraints, enabling the model to learn efficiently and robustly within complex high-dimensional feature spaces. In practice, we set the weights as \(\lambda_{\mathrm{inv}} = 0.01\) and \(\lambda_{\mathrm{cos}} = 0.1\) to appropriately balance the influence of each term, ensuring the model achieves not only superior fusion performance but also stable mappings and consistent semantic representations.

\section{Experiments}

\subsection{Experimental Settings}

\subsubsection{Dataset and Evaluation.}
We conduct experiments on three widely used hyperspectral datasets: Chikusei~\cite{Chikusei}, PaviaU, and Houston. The Chikusei image ($2517 \times 2335$, 128 bands) is split by using a $1000 \times 2000$ top-left region for training and the remaining area, divided into $680 \times 680$ patches, for testing. PaviaU ($610 \times 340$, 103 bands) uses the top $340 \times 340$ region for testing and the rest for training. For Houston ($349 \times 1905$, 144 bands), the left $349 \times 349$ area is used for testing. Following~\cite{simData}, HRMSI are simulated, and LRHSI are obtained via Gaussian blur (kernel size $3 \times 3$, std 0.5) and downsampling. Training samples are cropped into $64 \times 64$ ground-truth patches paired with corresponding LRHSI ($16 \times 16$) and HRMSI patches.

\subsubsection{Metrics.} For quantitative evaluation, we adopt four commonly-used metrics: Peak Signal-to-Noise Ratio (PSNR), Spectral Angle Mapper (SAM)~\cite{sam}, Relative Global Dimensional Synthesis Error (ERGAS)~\cite{ERGAS}, and Structural Similarity Index (SSIM)~\cite{SSIM}.

\subsubsection{Training Details.}

Our model is implemented in PyTorch and trained on an NVIDIA A30 GPU. The hidden dimension $D$ of PIF-Net is set to 64, and the number of feature extraction blocks $L$ is 4. We adopt the AdamW optimizer~\cite{adam} with an initial learning rate of $1 \times 10^{-4}$, which is halved every 200 epochs. The batch size is 8, and the total training spans 500 epochs. For all comparison methods, losses follow their official default implementations.

\subsection{Ablation Study}

\subsubsection{Effects of Ill-Posed Residual Priori. }
\begin{figure*}[htb]
    \centering
    \includegraphics[width=1\textwidth]{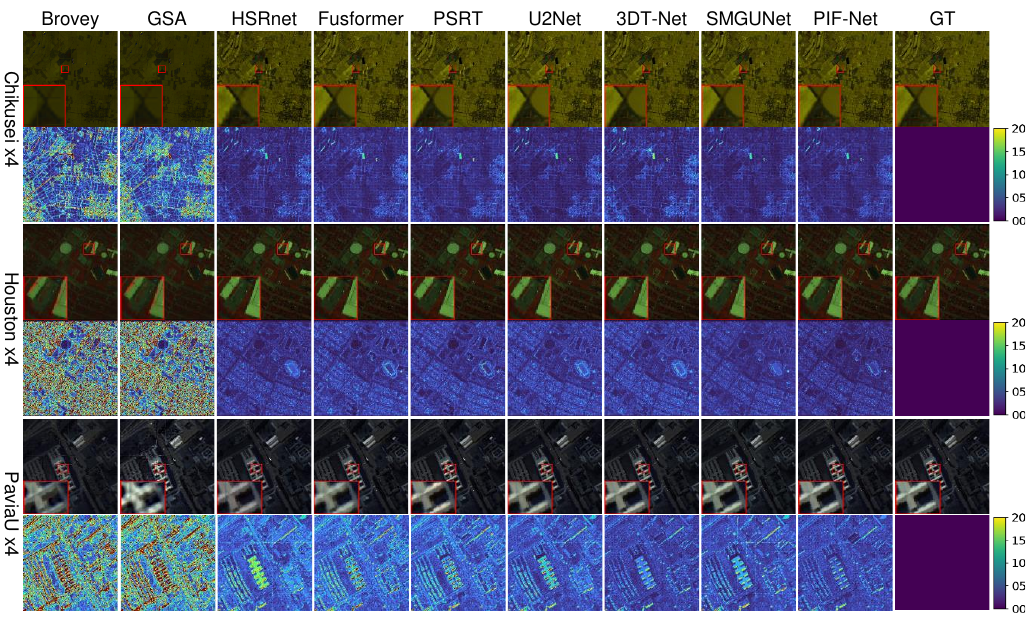}
	\caption{The pseudo-color images, corresponding $\times$4 super-resolution results, and SAM error maps generated by all comparative models on three datasets: (1) the sixth test area of the Chikusei dataset (rows 1–2, RGB bands: R=101, G=40, B=10), (2) a test area from the PaviaU dataset (rows 3–4, RGB bands: R=20, G=30, B=40), and (3) a test area from the Houston dataset (rows 5–6, RGB bands: R=10, G=76, B=2).}
    \label{fig:main_dis}
\end{figure*}

To investigate the influence of the ill-posed residual prior on model performance, we introduce a weighting parameter $\beta$ to control the strength of this prior. A series of ablation experiments were conducted on the PaviaU $\times 4$ dataset to evaluate the effect of varying $\beta$ values. The results, presented in Table~\ref{tab:prior}, reveal a clear trend in model performance as $\beta$ changes. Specifically, the ill-posed residual prior significantly contributes to improving both model stability and fusion accuracy. The best performance is achieved when $\beta = 1$. In contrast, setting $\beta$ to 0.4 or 0.8 results in no notable improvement and may even slightly degrade performance. Moreover, when $\beta = 0$, indicating the model does not utilize the residual prior, performance is at its lowest, further underscoring the critical role of this prior.

\begin{table}[htbp]
\setlength\tabcolsep{3pt} 
\centering
\resizebox{1\linewidth}{!}{
\begin{tabular}{ccccc}
\toprule
\(\beta\) & 0 & 0.4 & 0.8 & 1 \\
\midrule
PSNR & 36.9541 dB & 38.0012 dB & 38.9129 dB & \textbf{39.8246} dB \\
\bottomrule
\end{tabular}
}
\caption{Impact of varying $\beta$ values on PSNR performance on the PaviaU $\times 4$ dataset, demonstrating how the weighting of the ill-posed residual prior influences fusion quality.  The \textbf{best} results are in bold.}
\label{tab:prior}
\end{table}

\subsubsection{Effects of the Invertible Mamba Block.}
To assess the impact of the \textit{Invertible Mamba Block} on fusion performance, we performed an ablation study on the PaviaU $\times 4$ dataset, comparing the model's performance with and without this block. As shown in Table \ref{tab:ca_effect_mamba}, incorporating the Mamba Block leads to a notable improvement in fusion quality. Specifically, the PSNR increases from 37.1781 dB to 39.8246 dB, while SSIM improves from 0.9637 to 0.9845. These results highlight the effectiveness of the Mamba Block in enhancing feature representation and boosting the fidelity and quality of the fused images.

\begin{table}[htbp]
\centering
\resizebox{1\linewidth}{!}{
\begin{tabular}{lcc}
\toprule
Structure & w/o Mamba Block & w/ Mamba Block \\
\midrule
PSNR / SSIM & 37.1781 dB / 0.9637 & \textbf{39.8246 dB / 0.9845} \\
\bottomrule
\end{tabular}
}
\caption{Performance impact of the \textit{Invertible Mamba Block} on the PaviaU $\times 4$ dataset. The \textbf{best} results are in bold. }
\label{tab:ca_effect_mamba}
\end{table}

\subsubsection{Effects of FAM-LoRA.}
We also investigate the influence of the FAM-LoRA module using a similar ablation study on the PaviaU $\times 4$ dataset. As reported in Table \ref{tab:ca_effect_fam}, the inclusion of FAM-LoRA considerably enhances the model’s accuracy. Without this module, the PSNR and SSIM are 36.9211 dB and 0.9607, respectively. When FAM-LoRA is integrated, these metrics rise substantially to 39.8246 dB and 0.9845. This demonstrates that FAM-LoRA effectively facilitates feature aggregation, leading to superior fine-detail fusion and overall image quality.
\begin{table}[htbp]
\centering
\setlength\tabcolsep{6pt} 
\resizebox{1\linewidth}{!}{
\begin{tabular}{lcc}
\toprule
Structure & w/o FAM-LoRA & w/ FAM-LoRA \\
\midrule
PSNR / SSIM & 36.9211 dB / 0.9607 & \textbf{39.8246 dB / 0.9845} \\
\bottomrule
\end{tabular}}
\caption{Performance impact of the FAM-LoRA module on the PaviaU $\times 4$ dataset. The \textbf{best} results are in bold.}
\label{tab:ca_effect_fam}
\end{table}

\subsubsection{Ablation Study on Loss Components.}
To evaluate the impact of the invertibility regularization term \(\mathcal{L}_{inv}\) and the cosine similarity loss \(\mathcal{L}_{cos}\) on the overall loss function, we conducted an ablation study. The fusion loss \(\mathcal{L}_1\) is always included as the fundamental loss. Table~\ref{tab:ablation_loss} shows the PSNR and SSIM results of the model on the benchmark dataset under different loss combinations. When only \(\mathcal{L}_1\) is used, the model achieves a PSNR of 37.4521 dB and an SSIM of 0.9187. Adding \(\mathcal{L}_{inv}\) significantly improves performance, with PSNR reaching 38.6794 dB and SSIM 0.9342. Adding \(\mathcal{L}_{cos}\) also enhances the results, with PSNR of 38.1158 dB and SSIM of 0.9278. The best performance is obtained when both \(\mathcal{L}_{inv}\) and \(\mathcal{L}_{cos}\) are included, achieving a PSNR of 39.8246 dB and SSIM of 0.9845. These results indicate that the two auxiliary loss terms effectively improve fusion accuracy and semantic consistency, confirming their important roles in PIF-Net.
\begin{table}[ht]
\centering
\setlength\tabcolsep{11.5pt} 

\begin{tabular}{ccccc}
\toprule
$\mathcal{L}_1$ & $\mathcal{L}_{inv}$ & $\mathcal{L}_{cos}$ & PSNR (dB) & SSIM \\
\midrule
\ding{51} & \ding{55} & \ding{55} & 37.4521 & 0.9187 \\
\ding{51} & \ding{51} & \ding{55} & 38.6794 & 0.9342 \\
\ding{51} & \ding{55} & \ding{51} & 38.1158 & 0.9278 \\
\ding{51} & \ding{51} & \ding{51} & \textbf{39.8246} & \textbf{0.9845} \\
\bottomrule
\end{tabular}
\caption{Ablation study on loss components (\(\mathcal{L}_1\) loss always included) conducted on the PaviaU \(\times 4\) dataset. The \textbf{best} results are in bold.}
\label{tab:ablation_loss}
\end{table}

\subsection{Comparison with State-of-the-Art Methods}
\subsubsection{Quantitative Results.}
We comprehensively evaluate PIF-Net against several SOTA methods, with results in Table~\ref{tab:main} showing its clear superiority across all metrics. PIF-Net consistently achieves the highest PSNR and SSIM and the lowest SAM and ERGAS across datasets and scales. Notably, on challenging $\times4$ and $\times8$ tasks, PIF-Net significantly outperforms competitors—for instance, on PaviaU $\times4$, it surpasses the second-best by 0.9045 dB PSNR, and on Chikusei $\times8$, it sets a new benchmark with 50.0124 dB PSNR. These results validate that the integration of the Invertible Mamba Block and FAM-LoRA effectively captures spectral-spatial correlations and preserves high-frequency details, empowering PIF-Net’s superior performance, especially at large scaling factors

\subsubsection{Visual Comparison.}
\begin{figure}[!h]
    \centering
    \includegraphics[width=1\linewidth, trim=10 10 0 0, clip]{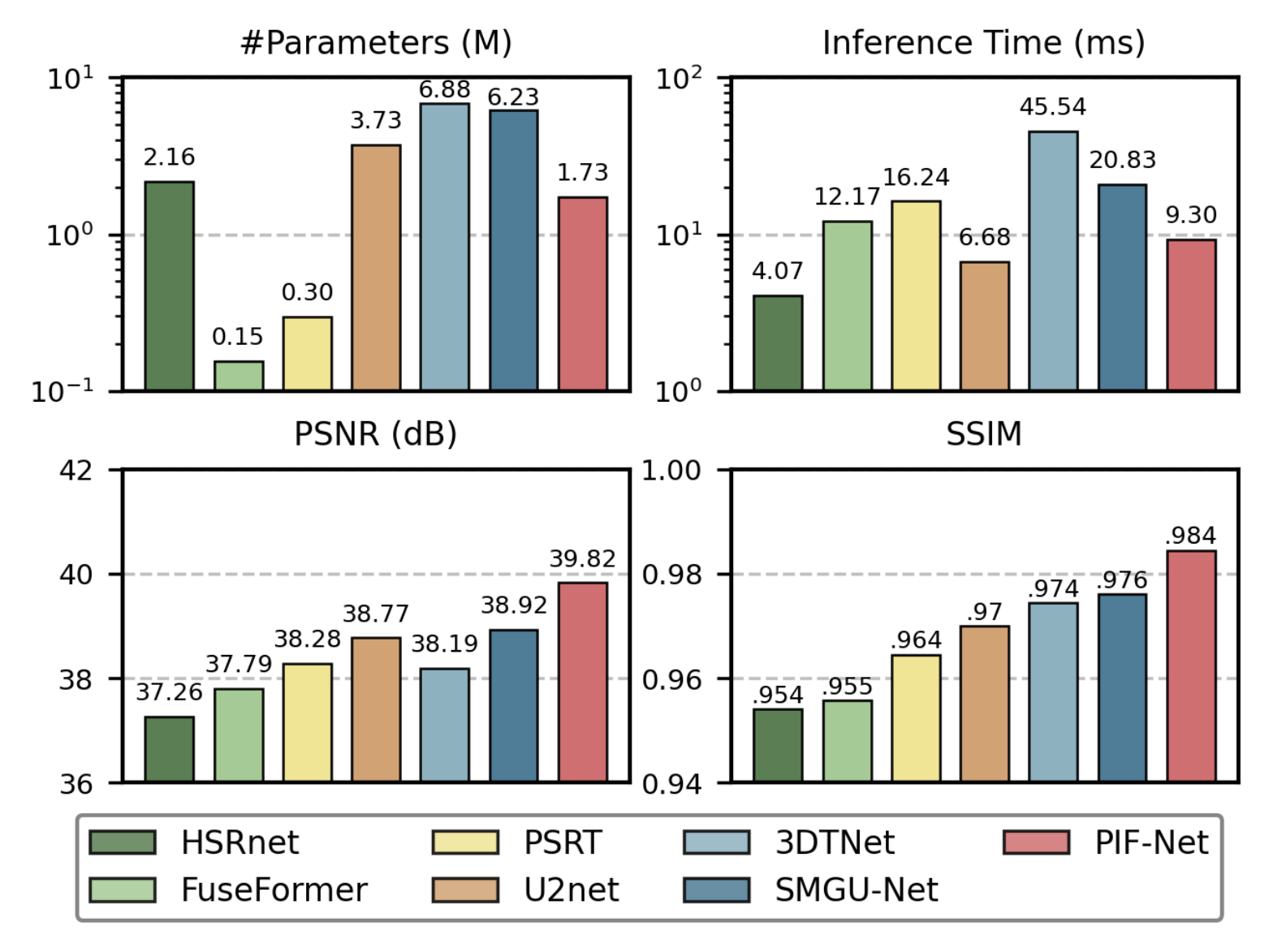}
 \caption{Comparison of model efficiency on PaviaU for $\times4$ super-resolution. Both the number of parameters (in millions) and inference latency (in milliseconds, measured on an NVIDIA A30 GPU) are displayed on logarithmic scales.}
    \label{fig:efficiency_comparison}
\end{figure}

We evaluate the computational efficiency of the models on the PaviaU dataset for the $\times$4 super-resolution task, measuring inference latency on an NVIDIA A30 GPU under identical implementation settings. As shown in Figure~\ref{fig:efficiency_comparison}, both the number of parameters and the inference latency are presented on a logarithmic scale for clearer comparison across models. PIF-Net achieves a favorable trade-off among model size, inference speed, and reconstruction accuracy. With only 1.73 million parameters and a latency of 9.3 ms, PIF-Net achieves the highest PSNR and SSIM among all compared methods, demonstrating superior efficiency and strong potential for real-time applications.

\subsubsection{Efficiency Comparison.}  
We evaluate the computational efficiency of all competing methods on the PaviaU dataset for the $\times4$ super-resolution task, measuring inference latency on an NVIDIA A30 GPU under identical implementation and batch settings. As illustrated in Figure~\ref{fig:efficiency_comparison}, both model size (in millions of parameters) and inference latency (in milliseconds) are displayed on logarithmic scales to better highlight relative differences. PIF-Net achieves a compelling balance among model complexity, speed, and reconstruction fidelity. With only 1.73M parameters and a latency of 9.3 ms, it outperforms all other methods in terms of both PSNR and SSIM, demonstrating SOTA accuracy and high computational efficiency, which makes it well suited for practical and real-time fusion scenarios.

\section{Conclusion}
This paper proposes PIF-Net, an innovative MHIF framework. Guided by ill-posed priors, the framework employs a dual-branch architecture that integrates HWT with Invertible Mamba Blocks, achieving high-fidelity information transformation and stable transmission, while ensuring the invertibility of feature transformations and the stability of gradient flow. Meanwhile, by combining a spatial-domain fusion strategy based on low-frequency priors with a lightweight LoRA fusion module, it effectively enhances the dynamic calibration of spatial details and spectral features. Furthermore,
we innovatively design a guided feature consistency loss that improves semantic alignment and representation across domains. Extensive experimental results demonstrate that PIF-Net achieves SOTA performance on three public datasets.
\section{Acknowledgments}
This research was supported by the National Key Research and Development Program of China (No. 2023YFB4502304). 
\bibliography{main}
\end{document}